\documentclass{article}
\usepackage{caption}
\usepackage{microtype}
\usepackage{graphicx}
\usepackage{subfigure}
\usepackage{booktabs} 
\usepackage{multirow} 
\usepackage{adjustbox} 
\usepackage{arydshln}
\usepackage{bm}

\usepackage{hyperref}

\usepackage[accepted]{icml2024}

\usepackage{amsmath}
\usepackage{amssymb}
\usepackage{mathtools}
\usepackage{amsthm}

\usepackage[capitalize,noabbrev]{cleveref}

\theoremstyle{plain}
\newtheorem{theorem}{Theorem}[section]

\theoremstyle{definition}

\theoremstyle{remark}

\usepackage[textsize=tiny]{todonotes}

\icmltitlerunning{Towards Independence Criterion in Machine Unlearning of Features and Labels}

\begin{document}

\twocolumn[
\icmltitle{Towards Independence Criterion in Machine Unlearning of Features and Labels}




\begin{icmlauthorlist}
\icmlauthor{Ling Han}{YaleCS,YaleM}
\icmlauthor{Nanqing Luo}{PSU}
\icmlauthor{Hao Huang}{WHUCS}
\icmlauthor{Jing Chen}{WHUCE}
\icmlauthor{Mary-Anne Hartley}{YaleM,EPFL}
\end{icmlauthorlist}

\icmlaffiliation{YaleCS}{Department of Computer Science, Yale University}
\icmlaffiliation{WHUCS}{School of Computer Science, Wuhan University}
\icmlaffiliation{WHUCE}{School of Cyber Science and Engineering, Wuhan University}
\icmlaffiliation{PSU}{College of Information Sciences and Technology, Penn State University}
\icmlaffiliation{YaleM}{School of Medicine, Yale University}
\icmlaffiliation{EPFL}{School of Computer and Communication Sciences, EPFL}

\icmlcorrespondingauthor{}{\{l.han, mary-anne.hartley\}@yale.edu, nkl5280@psu.edu, \{haohuang, chenjing\}@whu.edu.cn}

\icmlkeywords{Machine Learning, ICML}

\vskip 0.3in
]



\printAffiliationsAndNotice{}  

\begin{abstract}
This work delves into the complexities of machine unlearning in the face of distributional shifts, particularly focusing on the challenges posed by non-uniform feature and label removal. With the advent of regulations like the GDPR emphasizing data privacy and the right to be forgotten, machine learning models face the daunting task of unlearning sensitive information without compromising their integrity or performance. Our research introduces a novel approach that leverages influence functions and principles of distributional independence to address these challenges. By proposing a comprehensive framework for machine unlearning, we aim to ensure privacy protection while maintaining model performance and adaptability across varying distributions. Our method not only facilitates efficient data removal but also dynamically adjusts the model to preserve its generalization capabilities. Through extensive experimentation, we demonstrate the efficacy of our approach in scenarios characterized by significant distributional shifts, making substantial contributions to the field of machine unlearning. This research paves the way for developing more resilient and adaptable unlearning techniques, ensuring models remain robust and accurate in the dynamic landscape of data privacy and machine learning.
\end{abstract}

\section{Introduction}
Machine learning models, with their remarkable ability to discern patterns, might inadvertently capture sensitive information from their training data, raising genuine privacy concerns. While these models are designed to generalize from data, there exists a possibility that they could unintentionally reveal sensitive details upon closer examination. The European General Data Protection Regulation (GDPR) \cite{regulation2016regulation} emphasizes the "right to be forgotten", which primarily pertains to the erasure of personal data from databases. Translating this principle to machine learning models introduces notable challenges, especially when the requests for data removal are more granular than individual data points and extend to specific features or labels, thereby complicating the unlearning process \cite{warnecke2021machine}.

Delving deeper into the unlearning process, one observes that data removal requests, especially those concerning specific features or labels, often pertain to data elements that aren't uniformly distributed within the training set \citep{erdemir2021adversarial, li2023critical}. Such non-uniform deletions can potentially induce noticeable shifts in the data distribution \citep{yonekawa2022riden, agarwal2022minimax}. To illustrate this, imagine a situation where a credit company's clients seek to erase their voluntarily disclosed records. It is plausible that records associated with certain behaviors, perhaps those deemed "at fault" could be more commonly selected for deletion.

While the broader challenges of machine unlearning in distributional shifts have been covered in various contexts, such as accuracy \citep{bourtoule2021machine}, timeliness \citep{cao2015towards}, and privacy \citep{chen2021machine}, the nuanced challenges arising from unlearning features and labels in scenarios of non-uniform removal warrant a closer look. With this research, we aim to navigate this intricate terrain, focusing on the challenges and exploring potential strategies for effective feature and label unlearning.

Several strategies have been proposed for machine unlearning in the context of distributional shifts. The most obvious approach is simply retraining the model with an adjusted dataset that omits the targeted data points and features\citep{nguyen2022survey}. Retraining has the potential to address core concerns of accuracy and privacy in the face of distributional shifts \citep{yan2022arcane}. However, its practicality is sometimes questioned due to issues related to timeliness \citep{foster2023fast}. The inherent computational demands, combined with the need for original training data storage, can make this method less suitable for applications requiring real-time responsiveness \citep{schelter2021hedgecut}. Another set of strategies, known as indistinguishability-based methods attempts to obscure the weights that were cached during training,\citep{wu2020deltagrad, golatkar2020eternal, guo2019certified, sekhari2021remember}. These methods have been compared with differential privacy and are particularly efficient for localized unlearning requests, which characterized by a concentration in data structure or associative relationships. Yet, their performance can be less consistent in scenarios where unlearning requests are more widespread without being constrained by structural or local associations. A notable concern with these methods is the potential compromise of data set integrity due to fragmentation \citep{nguyen2022survey}, which can be especially challenging for datasets like graph data that depend on a holistic structure \citep{chen2022graph}.

When it comes to unlearning specific features and labels, there is growing interest in approximation methods rooted in robust statistics \citep{warnecke2021machine, wu2023gif, zhang2023recommendation}. While these methods have demonstrated effectiveness in typical settings, they can face challenges, such as model distortions, in environments experiencing distributional shifts \citep{giraud2021introduction, jones2018time}. These distortions highlight the importance of developing more resilient unlearning techniques.

In light of these observations, our research seeks to understand and address these complexities, with the goal of proposing strategies that harmonize accuracy, privacy, timeliness, and computational efficiency, when confronted with distributional shifts in the machine unlearning process. 


In response to these challenges, our work introduces a novel approach to machine unlearning that specifically addresses the intricacies of distributional shifts caused by non-uniform feature and label removal. Our method, grounded in the principles of influence functions and distributional independence, seeks to refine the unlearning process, ensuring both privacy protection and model integrity in the face of such shifts. By leveraging the influence function, we propose a mechanism that not only facilitates efficient data removal but also dynamically adjusts the model to maintain its performance and generalization capabilities across varying distributions.

Our research makes several pivotal contributions to the field of machine unlearning, particularly in scenarios characterized by distributional shifts:
\begin{itemize}
    \item We introduce a comprehensive framework for machine unlearning that effectively addresses the challenges posed by non-uniform feature and label removal. This framework is designed to ensure privacy protection while minimizing the impact on model performance and distributional integrity.
    \item Through the application of influence functions, we develop a method that allows for the precise estimation and removal of data points' influence on the model, thereby facilitating a more nuanced approach to unlearning.
    \item Our extensive experiments demonstrate the efficacy of our approach, showcasing its superiority in maintaining model accuracy and integrity, even in the presence of significant distributional shifts.
\end{itemize}

\section{Preliminary}


\newcommand{\mD}{\mathcal{D}}
\newcommand{\mX}{\mathcal{X}}
\newcommand{\mI}{\mathcal{I}}
\newcommand{\mF}{\mathcal{F}}
\newcommand{\vf}{\mathbf{f}}  

\begin{table*}[htbp]
\centering
\caption{\textbf{List of Notations and Symbols}}
\bgroup
\def\arraystretch{1.5}
\begin{tabular}{p{2in}p{3.75in}}
\hline Notations and Symbols & Description \\
\hline $\displaystyle \mD =\{z_1, z_2,\cdots, z_k\}$ & Training Data Set with $k$ Data Points\\
$\displaystyle \Delta\mD$ & Data Perturbation in Unlearning Process\\
$\displaystyle \mX=\{x_1, x_2,\cdots, x_m\}$ & The $m$ Features in Training Data Set\\
$\displaystyle z_i=(x^i_1,x^i_2,\cdots,x^i_m, Y_i)$ & The $i$-th Training Data Point Consist of Features and Labels \\
$\displaystyle \mathcal{F}(x_i)$ & The Distribution of the $i$-th Feature in Training Data Set\\
$\displaystyle \mathcal{F}(\mX)$ & The Distribution of the Features in Training Data Set\\
$\displaystyle \mathcal{F}(\mD)$ & The Distribution of Features and Labels in Training Data Set $\mD$\\
$\displaystyle \theta ; \theta^*$ & Model Parameters; Updated Model Parameters \\
$\displaystyle \bm{f}(\theta;\mD)$ & Model with Parameters $\theta$ Trained on Data Set $\mD$ \\
$\displaystyle \bm{f}(\theta^*;\mD\backslash\Delta\mD)$ & Unlearned Model with Parameters $\theta$ Trained on Data Set $\mD\backslash\Delta\mD$ \\
$\displaystyle \mI\mF(z_i;\bm{f}(\theta;\mD))$ & The Influence of Data Point $z_i$ to the Model $\bm{f}(\theta;\mD)$\\
$\displaystyle \bm{l}(z_i;\bm{f}(\theta;\mD))$ & Loss Perturbation to the Model $\bm{f}(\theta;D)$ Caused by Data Point $z_i$\\
\hline
\end{tabular}
\egroup
\vspace{0.25cm}
\end{table*}

Central to our exploration is the challenge of detecting distributional shifts during the machine unlearning process. We consider a training dataset \(\mathcal{D} = \{z_1, z_2, \ldots, z_k\}\), where each data point \(z_i = (X_i, Y_i)\) consists of an \(m\)-dimensional feature vector \(X_i = \{x^i_1, x^i_2, \ldots, x^i_m\}\) and a label \(Y_i\). The objective in distributional shift detection is to train a model \(\mathbf{f}(\theta; \mathcal{D})\) parameterized by \(\theta\), which minimizes the loss \(\mathbf{l}(z_i; \mathbf{f}(\theta; \mathcal{D}))\) between the predicted labels and the true labels \(Y_i\) for each data point \(z_i\) in \(\mathcal{D}\).

Upon receiving an unlearning request, a subset \(\Delta \mathcal{D}\) of the data points, consisting of both features and labels, is identified for removal from \(\mathcal{D}\). The goal of machine unlearning, in the presence of distributional shifts, is to establish a mechanism \(M\) that, given \(\mathcal{D}\), \(\mathbf{f}(\theta; \mathcal{D})\), and \(\Delta \mathcal{D}\), produces an updated model \(\mathbf{f}(\theta^*; \mathcal{D} \setminus \Delta \mathcal{D})\) with parameters \(\theta^*\). This model should closely emulate the performance of a model retrained from scratch on \(\mathcal{D} \setminus \Delta \mathcal{D}\), even when there are significant divergences between the distributions \(\mathcal{F}(\mathcal{D})\) and \(\mathcal{F}(\Delta \mathcal{D})\). A machine unlearning method responsive to distributional changes should satisfy the following criteria:

\begin{enumerate}
  \item \textbf{Comprehensive Unlearning:} The primary objective of machine unlearning is the complete removal of target data from the training dataset. This process should not only ensure privacy protection but also thoroughly eliminate the direct and latent influences of the data on the model.

  \item \textbf{Adaptability:} Following the excision of specific data, the model must demonstrate a robust fit to the remaining dataset, accommodating the new data distribution and information content.

  \item \textbf{Efficiency:} The unlearning procedure should be efficient, circumventing the extensive time requirements synonymous with retraining models from scratch.

  \item \textbf{Generalization:} The unlearning approach should maintain its efficacy across various models and distributional changes, rather than being confined to specific distribution alterations or particular model architectures.
\end{enumerate}

The foundational concept underpinning our approach to machine unlearning with influence functions \citep{warnecke2021machine} in the presence of distributional shifts. We draw upon the principles of the Information Bottleneck (IB) \citep{tishby2000information, goldfeld2020information} method and the Hilbert-Schmidt Independence Criterion (HSIC) \citep{gretton2005measuring, ma2020hsic} to develop a framework that effectively addresses the challenges posed by non-uniform feature and label distribution shifts in unlearning tasks.

\textbf{Mutual Information:} Mutual information (MI) quantifies the shared information between two variables, crucial for measuring the dependency between input features \(X\) and output labels \(Y\) in machine learning. The objective is to maximize \(I(X;Y)\), represented as:
\begin{equation}
\max_{\theta} \; I(X;Y),
\end{equation}
where \( \theta \) denotes the model parameters. This maximization ensures the model captures pivotal relationships within the data, enhancing prediction accuracy and generalization across diverse data distributions.

\textbf{Hilbert-Schmidt Independence Criterion (HSIC):} HSIC serves as a non-parametric measure of independence between random variables. The empirical expression of HSIC is given by:
\begin{equation}
nHSIC[X,Y] = tr(\tilde{K}_X \tilde{K}_Y)
\end{equation}
where \( \tilde{K}_X \) and \( \tilde{K}_Y \) are the centered kernel matrices for the input features \( X \) and labels \( Y \), respectively.

\section{DUI: Distributional Unlearning with Independence Criterion} 
Let us consider a more nuanced scenario: how can we represent the impact of altering a training data point, specifically in terms of measuring its influence on distributional changes when we aim to unlearn that particular point? 

\subsection{Distribution Difference and Independence Criterion}
To provide a more tangible expression of the impact we aim to measure on distributional changes, we first formalize the problem at hand. For a training data point \(z = (X_i, Y_i)\), consider the perturbation empirical risk minimizer \(\hat{\theta}_{-z} - \theta\) and its distribution shift \(\mathcal{D}\mathcal{S}(z; \bm{f}(\hat{\theta}; \mathcal{D}))\). 

To assess potential shifts in the conditional probability distribution \( P(Y \mid X) \), we first quantify the correlation between features \( X \) and labels \( Y \). This method is intended to detect significant changes in the dependency relationship, indicative of variations in \( P(Y \mid X) \). Mutual information (MI) is a commonly used criterion and can be estimated as:
\begin{equation}
    M I\left(X, Y\right)=\hat{P}\left(X, Y\right) \log \frac{\hat{P}\left(X, Y\right)}{\hat{P}\left(X\right) \hat{P}\left(Y\right)}
\end{equation}
Considering that the accuracy of predicted probabilities for conditional distribution represents the distributional differences, we attempt to use independence to signify distributional changes, denoted as $\Delta\hat{F}(Z; \theta) = \Delta MI(X, \hat{Y}; \theta)$. Furthermore, based on the definition of loss in machine learning, we express the component of distributional independence in the loss as
\begin{equation}
\label{eq2}
    L_F = DIST[MI(X,Y), MI(X, \hat{Y}; \theta)]
\end{equation}
Here $DIST$ represents distance in Euclidean space \citep{ribeiro2016should}. And \(\hat{Y};\theta\) represents the prediction set of the model with parameter \(\theta\). 

\subsection{Relation to Unlearning} 

In the context of unlearning, scholars \citep{warnecke2021machine, wu2023gif} employ the influence function to meticulously analyze the impact of specific data points selected for removal. 
\begin{theorem}
\label{theo:IF}
Assume we have a unlearning request to remove a subset \(\Delta \mathcal{D}\) from \(\mathcal{D}\), then the parameters will be adjusted during unlearning for $\Delta\theta = H^{-1}_{\theta_0} \nabla_{\theta_0} \displaystyle \mI\mF(z\in\Delta \mathcal{D};\vf(\theta_0;\mD)) $. Where as $\mI\mF(z_i\in\Delta \mathcal{D};\bm{f}(\theta_0;\mD))= \Sigma_{z \in \Delta D} \mI\mF(z;\bm{f}(\theta_0;\mD))$.
\end{theorem}
Building upon the Theorem.\ref{theo:IF}, it is important to note that traditional approaches employing the influence function often rely on the original loss function of the model as the measure of influence. Specifically, the influence of a data point \(z\) in the subset \(\Delta \mathcal{D}\) to be unlearned is quantified as:
\begin{align}
\mI\mF(z \in \Delta \mathcal{D}; \vf(\theta_0; \mD)) &= \bm{l}(z \in \Delta \mathcal{D}; \bm{f}(\theta_0; \mD)) \nonumber \\
&= \sum_{z \in \Delta \mD} \bm{l}(z; \bm{f}(\theta_0; \mD))
\end{align}
where \(\bm{l}(z; \bm{f}(\theta_0; \mD))\) represents the loss associated with the data point \(z\).

In our endeavor to redefine the influence of a data point, we aim to more accurately reflect its impact on the distributional changes represented by the model. In the early studies of neural networks, regularization \citep{hornik1989multilayer} was introduced as a means to prevent overfitting. A loss function incorporating regularization can be expressed as \( L_{\text{total}} = L_{\text{origin}} + \lambda \cdot R(\theta) \), where \( L_{\text{origin}} \) is the original loss and \( R(\theta) \) represents the regularization term.

Further, regularization enhances the model's generalization ability by optimizing the representation of high-dimensional distributions encapsulated by the model. Inspired by this concept, we propose to represent the model's distributional shift through the regularization term. Accordingly, we define the loss function with an added focus on distributional changes, as follows:
\begin{equation}
   L_{\text{total}} = L_{\text{origin}} + \lambda \Delta P(\mathcal{D}, \mathcal{D}\backslash\Delta\mathcal{D})
\end{equation}
where \( \Delta P(\mathcal{D}, \mathcal{D}\backslash\Delta\mathcal{D}) \) quantifies the distributional shift due to the unlearning of data points in \( \Delta \mathcal{D} \). This formulation allows us to capture the essence of the distributional changes in the model's learning process, thereby providing a more holistic view of the influence exerted by the data points being unlearned.

We represent the impact of distributional changes using mutual information as follows:
\begin{flalign}
    &\Delta P(\mathcal{D}, \mathcal{D} \backslash \Delta \mathcal{D})
    = DIST\left(\text{MI}(X_{\mathcal{D}},Y_{\mathcal{D}}),\text{MI}(X_{\mathcal{D}}, \hat{Y}_{\theta})\right)\notag\\ 
    & -DIST\left(\text{MI}(X_{\mathcal{D} \backslash \Delta \mathcal{D}},Y_{\mathcal{D} \backslash \Delta \mathcal{D}}) ,\text{MI}(X_{\mathcal{D} \backslash \Delta \mathcal{D}}, \hat{Y}_{\theta})\right)
\end{flalign}
Where \( \text{MI}(X_{\mathcal{D}},\hat{Y}_{\theta}) \) denotes the mutual information between features \( X \) and original model prediction \(\hat{Y}_{\theta_0}\) in the original dataset \( \mathcal{D} \), and \( \text{MI}(X_{\mathcal{D} \backslash \Delta \mathcal{D}}, \hat{Y}_{\theta}) \) represents the mutual information in the dataset after the removal of \( \Delta \mathcal{D} \). The function \( \text{DIST} \) quantifies the difference in mutual information, thereby measuring the shift in the data distribution due to unlearning.


\subsection{Adjust Parameters for Distribution Shift}
Further, based on the influence function, we derive the following relationship:
\begin{align}
&\sum_{z_i \in \mathcal{D} \backslash \Delta \mathcal{D}} \mI\mF\left(z;\bm{f}\left(\theta;D\right)\right) \notag\\ &= \sum_{z_i \in \mathcal{D}} \mI\mF\left(z;\bm{f}\left(\theta;D\right)\right) - \sum_{z_i \in \Delta \mathcal{D}} \mI\mF\left(z;\bm{f}\left(\theta;D\right)\right)
\end{align}
The aforementioned relationship, derived through the analysis of each data point's influence, proposes a connection in the unlearning process. However, it does not incorporate the impact of distributional changes. In the approximation process, the influence function does not sufficiently adjust the parameters during updates to accurately reflect distributional shifts. 

To approximate effects of distributional shifts, we redefine the parameters resulting from the perturbation as \(\hat{\theta}_{\epsilon;-z} \stackrel{\text{def}}{=} \arg\min_{\theta \in \Theta} \Delta F(Z; \theta) + \epsilon \Delta F(z; \theta)\). Therefore, \(\hat{\theta}_{\epsilon;z}\) is redefined as \(\hat{\theta}_{\epsilon;z} \stackrel{\text{def}}{=} \arg \min _{\theta \in \Theta} \frac{1}{n} \sum_{i=1}^n L_F(z_i, \theta) + \epsilon L_F(-z, \theta)\). Following the analogy and derivations presented in \citep{koh2017understanding}, we arrive at the following conclusion:
\begin{equation}
\label{eq3}
\left.\mathcal{I}_{\text {up,params }}(z) \stackrel{\text { def }}{=} \frac{d \hat{\theta}_{\epsilon, z}}{d \epsilon}\right|_{\epsilon=0}=-H_{\hat{\theta}}^{-1} \nabla_\theta L_F(z, \hat{\theta})
\end{equation}
\(\mathcal{I}_{\text {up,params }}(z)\) represents influence function\citep{cook1980characterizations} approximately evaluates the influence of upweighting $z$ on the parameters $\hat{\theta}$. Where as $H_{\hat{\theta}}$ is the Hessian matrix of model \(f (\theta;D)\). 

To accurately reflect distributional shifts, we propose to split the influence into two parts: one measuring the change in the model's loss and the other estimating the impact of data points on the distribution. Consequently, we introduce the following formulation to address this:
\begin{equation}
\hat{\theta}_\epsilon = \underset{\theta}{\arg \min} \left(L_{\text{origin}} + \epsilon \left(\Delta L_{(\Delta \mathcal{D})} + \lambda \Delta L_{\text{F}(\Delta \mathcal{D})}\right)\right),
\end{equation}
where \( L_{\text{origin}} \) is the original loss function, \( \Delta L_{(\Delta \mathcal{D})} \) represents the change in loss due to the removal of data points in \( \Delta \mathcal{D} \), and \( \Delta L_{\text{F}(\Delta \mathcal{D})} \) quantifies the influence of these data points on the distributional shift. This approach allows for a more nuanced and accurate adjustment of parameters, reflecting both the loss variation and the distributional impact of the unlearning process.


\subsection{Calculating Influence of Independence}

In this section, we introduce a novel method for unlearning unevenly distributed features from a learning model. To more accurately reflect the independent relationships between datasets, as opposed to mutual information, we employ the Hilbert-Schmidt Independence Criterion (HSIC) as an independence measure to capture the changes in the training data distribution. The adapted formulation is:
\begin{equation}
L_{\text{F}} = DIST\left(\text{nHSIC}(X_{\mathcal{D}},Y_{\mathcal{D}}),\text{nHSIC}(X_{\mathcal{D}}, \hat{Y}_{\theta})\right)
\label{eq:IB1}
\end{equation}
Where \( \Delta L_{\text{F}} \) quantifies the distributional shift in the model's learning process by comparing the HSIC between the original features \( X_{\mathcal{D}} \) and labels \( Y_{\mathcal{D}} \) against the HSIC between the features and the model's predicted labels \( \hat{Y}_{\theta} \).

Given the iterative updates of parameters in the presence of a Hessian matrix, the formula with an extra scaling coefficient \(\beta\) \citep{wu2023gif} for Hessian-Vector-Products (HVPs) is expressed by:
\begin{align}
    H_j^{-1} v &= v + H_{j-1}^{-1} v - \beta H_{\theta_0}\left[H_{j-1}^{-1} v\right], H_0^{-1} v = v \nonumber\\
    &=v+\left(I-\beta \nabla_\theta^2 L_F\left(z_i, \theta^*\right)\right) \tilde{H}_{j-1}^{-1} v , H_0^{-1} v =v \nonumber\\
    &=v+\left(I-\nabla_\theta^2 \lambda L_F\left(z_i, \theta^*\right)\right) \tilde{H}_{j-1}^{-1} v , H_0^{-1} v =v
\end{align}
Where \(H_j^{-1}\) denotes the inverse Hessian matrix at iteration \(j\), \(v\) represents an arbitrary vector, and \(\beta\) is a scalar denoting the learning rate. 

We now combine the original loss function with the distributional shift for parameter updates, leading to the following representation:

\begin{align}
&H_j^{-1} v = v+\left[I-\nabla_\theta^2 \left(L(z_i, \theta^*)+\lambda L_F(z_i, \theta^*)\right)\right] \tilde{H}_{j-1}^{-1} v, \nonumber \\ &H_0^{-1} v =v
\end{align}

This formulation integrates the original loss function with the distributional shift, effectively capturing the influence of data points on the model's learning process. 

Particularly, within \(L_F\), we introduce a hyper-parameter \(\alpha\) to balance the relationship between the two independence terms. We represent the Euclidean distance between these independence measures by directly subtracting them:

\begin{equation}
L_F = \text{nHSIC}(X_{\mathcal{D}},Y_{\mathcal{D}}) - \alpha \text{nHSIC}(X_{\mathcal{D}}, \hat{Y}_{\theta})
\end{equation}

Through this parameter update mechanism, we achieve distributional unlearning. The related experiments demonstrating the effectiveness of this approach are discussed in the following sections.


\section{Experiments}
To validate the Adaptability, Efficiency, and Generalization of the DUI method, we conduct a series of experiments aimed at addressing the following research questions (RQ):
\begin{itemize}
    \item \textbf{RQ1: Efficiency and Generalization.} How does DUI compare with existing unlearning methods in terms of running time and accuracy of the final results? Can DUI demonstrate efficiency and generalization superior to conventional approaches?
    
    \item \textbf{RQ2: Unlearning under Distributional Shifts.} In scenarios where there is a significant distributional change before and after unlearning, can DUI achieve better unlearning in terms of accuracy and effectiveness?
    
    \item \textbf{RQ3: Adaptability through Distributional Adjustment.} Is DUI capable of effectively adjusting the model's distribution to enhance adaptability, thereby fully modifying the model's representation of the distribution?
\end{itemize}
\begin{table}
\centering
\caption{\textbf{Statistics of Datasets}}
\begin{adjustbox}{max width=\textwidth}
\begin{tabular}{c|ccccc}
\hline Dataset & Type & Points & Edges & Features & Classes \\
\hline Cora & Graph & 2,708 & 5,429 & 1,433 & 7 \\
Citeseer & Graph & 3,327 & 4,732 & 3,703 & 6 \\
MNIST & Image & 70,000 & N/A & 70,000 & 10 \\
\hline
\end{tabular}
\end{adjustbox}
\label{T2}
\end{table}
\subsection{Experimental Setup}
\subsubsection{Datasets}
We evaluate the DUI method's adaptability and efficiency using datasets from both Euclidean and non-Euclidean domains: MNIST \cite{lecun1998mnist} for images, and Cora \cite{kipf2016semi} and Citeseer \cite{kipf2016semi} for graphs. These selections allow us to test DUI across image classification with CNNs and node classification with GNNs. MNIST features are based on image pixels, with labels for the depicted digits. Cora and Citeseer consist of publication nodes linked by citations, with features indicating keyword presence. Each dataset is divided into a training set (90\%) and a test set (10\%), ensuring a comprehensive evaluation framework. Dataset statistics are detailed in Table \ref{T2}.

\subsubsection{Models}
For different datasets, we test the DUI method on a Simple CNN \cite{le1994word} model in Euclidean space and on GCN \cite{kipf2016semi}, and GIN \cite{xu2018powerful} models in graph space.

\subsubsection{Evaluation Metrics}
We evaluate the performance of the DUI method using F1 Score and Runtime, comparing it with other approaches. The F1 Score measures the harmonic average of precision and recall, providing a balanced view of the model's utility. Runtime reflects the efficiency of the unlearning process, and the Brier Score offers an assessment of the accuracy of predicted probabilities, particularly in the context of distributional shifts.


\subsubsection{Unlearn Request}
To assess the adaptability and efficiency of the DUI method under different scenarios, we implement two distinct settings for unlearning requests. 

\textbf{Random Unlearning:} In the first setting, we randomly select features and labels to unlearn. This approach does not alter the overall distribution of features and labels, thereby simulating traditional unlearning scenarios where the data removal is arbitrary and does not follow a specific pattern.

\textbf{Top-k Strategy:} The second setting adopts a more targeted approach, known as the top-k strategy. Here, we unlearn the top \( k = n \times unlearn\_ ratio \) features and labels with the highest values, effectively removing them from the training dataset. This strategy is designed to mimic real-world scenarios where unlearning requests might lead to significant distributional changes, particularly in cases where certain features or labels are disproportionately represented.

\begin{figure}[ht]
\centering
\includegraphics[width=0.5\textwidth]{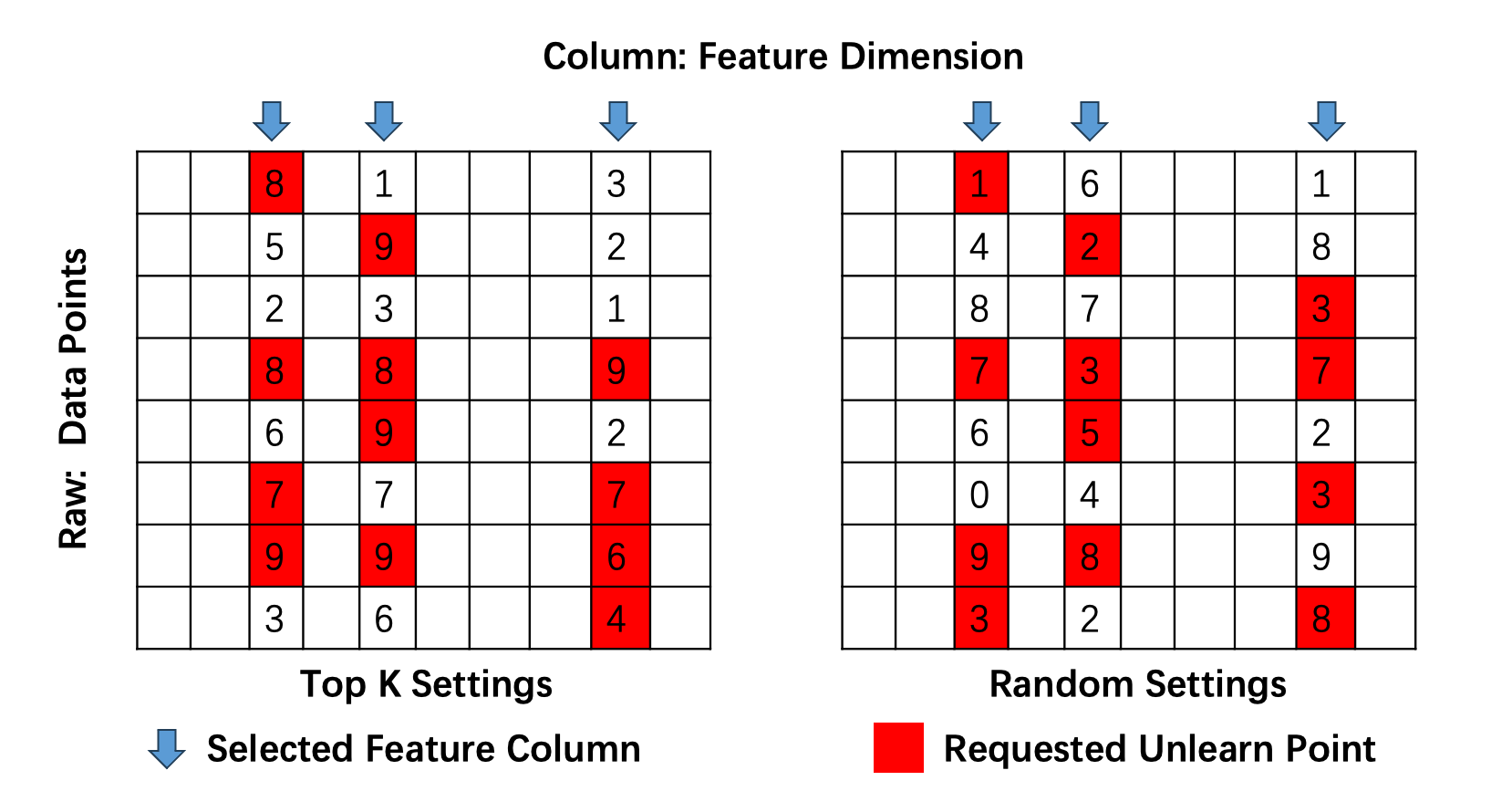}
\caption{Illustration of the unlearn request process in graph settings, showcasing the application of feature ratio and unlearn ratio metrics.}
\label{fig:unlearn_process}
\end{figure}

\textit{Graph Setting Considerations:} For graph datasets, we employ feature and unlearn ratios to guide the unlearning process. The feature ratio specifies the fraction of features affected, and the unlearn ratio determines the quantity of feature values removed within each group. In scenarios testing distributional shifts, such as the top-k setting, we selectively unlearn the top-k values per feature group, as dictated by the unlearn ratio. This method simulates realistic distributional changes, targeting specific data points for removal. Figure \ref{fig:unlearn_process} depicts this targeted unlearning strategy.

\begin{table*}
\centering
\caption{\textbf{Top K Settings: Comparison of F1 scores and running time (RT) for different methods for feature unlearning with $5\%$ to $20\%$ data points deleted from the original dataset.}}
\begin{adjustbox}{max width=\textwidth}
\begin{tabular}{cccccccc}
\hline 
\multicolumn{2}{c}{\textbf{Iteration}} & \multicolumn{6}{c}{\textbf{Dataset / Model}} \\
\hline 
\multirow{2}{*}{\textbf{Unlearn Ratio}} & \multirow{2}{*}{\textbf{Method}} & \multicolumn{2}{c}{\textbf{Cora / GIN}} & \multicolumn{2}{c}{\textbf{Citeseer / GCN}} & \multicolumn{2}{c}{\textbf{MNIST / Simple CNN}} \\
\cline{3-8} 
& & \textbf{F1 score} & \textbf{RT (second)} & \textbf{F1 score} & \textbf{RT (second)} & \textbf{F1 score} & \textbf{RT (second)} \\
\hline 
\multirow{4}{*}{0.05} & Retrain & $0.8057 \pm 0.0159$ & 8.31 & $0.7318 \pm 0.0096$ & 7.52 & $0.9587 \pm 0.0005$ & 198.96 \\ 
\cdashline{2-8}
& IF & $0.7738 \pm 0.0153$ & 0.48 & $0.7001 \pm 0.0228$ & 0.25 & $0.8978 \pm 0.0181$ & 4.38 \\
& GIF & $0.7828 \pm 0.0201$ & 0.50 & $0.7102 \pm 0.0143$ & 0.19 & N/A & N/A \\
& \textbf{DUI} & $\mathbf{0.7868 \pm 0.0189}$ & 0.99 & $\mathbf{0.7135 \pm 0.0158}$ & 0.37 & $\mathbf{0.9433 \pm 0.0176}$ & 8.75 \\
\hline
\multirow{4}{*}{0.075} & Retrain & $0.8051 \pm 0.0102$ & 8.32 & $0.7317 \pm 0.0043$ & 7.38 & $0.9574 \pm 0.0014$ & 192.37 \\
\cdashline{2-8}
& IF & $0.7739 \pm 0.0180$ & 0.72 & $0.7023 \pm 0.0126$ & 0.37 & $0.8974 \pm 0.0190$ & 7.08 \\
& GIF & $0.7804 \pm 0.0163$ & 0.73 & $0.7083 \pm 0.0127$ & 0.38 & N/A & N/A \\
& \textbf{DUI} & $\mathbf{0.7853 \pm 0.0181}$ & 1.47 & $\mathbf{0.7125 \pm 0.0167}$ & 0.69 & $\mathbf{0.9427 \pm 0.0353}$ & 12.97 \\
\hline
\multirow{4}{*}{0.1} & Retrain & $0.8048 \pm 0.0132$ & 8.18 & $0.7312 \pm 0.0087$ & 7.49 & $0.9543 \pm 0.0012$ & 189.32 \\
\cdashline{2-8}
& IF & $0.7730 \pm 0.0213$ & 0.98 & $0.7024 \pm 0.0235$ & 0.45 & $0.8932 \pm 0.0203$ & 8.34 \\
& GIF & $0.7797 \pm 0.0199$ & 0.82 & $0.7083 \pm 0.0185$ & 0.48 & N/A & N/A \\
& \textbf{DUI} & $\mathbf{0.7805 \pm 0.0203}$ & 1.71 & $\mathbf{0.7117 \pm 0.0157}$ & 0.87 & $\mathbf{0.9407 \pm 0.0283}$ & 17.83 \\
\hline
\multirow{4}{*}{0.2} & Retrain & $0.8016 \pm 0.0094$ & 8.27 & $0.7293 \pm 0.0049$ & 7.12 & $0.9509 \pm 0.0033$ & 185.45 \\
\cdashline{2-8}
& IF & $0.7712 \pm 0.0207$ & 1.06 & $0.7008 \pm 0.0235$ & 0.46 & $0.8919 \pm 0.0198$ & 9.88 \\
& GIF & $0.7764 \pm 0.0238$ & 1.08 & $0.7106 \pm 0.0185$ & 0.46 & N/A & N/A \\
& \textbf{DUI} & $\mathbf{0.7798 \pm 0.0198}$ & 1.97 & $\mathbf{0.7115 \pm 0.0142}$ & 0.98 & $\mathbf{0.9371 \pm 0.0304}$ & 20.03 \\
\hline
\end{tabular}
\end{adjustbox}
\end{table*}

\subsection{Efficiency and Adaptability Analysis (RQ1)}
Our investigation into RQ1 contrasts DUI with traditional retraining and other methods like Influence Function (IF) and Graph Influence Function (GIF) across various datasets and models, focusing on unlearning ratios from 5\% to 20\%. Our findings reveal:
    \textbf{Efficiency:} DUI markedly reduces the time required for unlearning, demonstrating significant efficiency improvements. For example, retraining times for graph datasets and MNIST at a 5\% unlearning ratio are substantially longer compared to DUI's more rapid processing, without sacrificing model utility.
    
    \textbf{Generalization:} DUI maintains competitive F1 scores across all datasets, showcasing its robustness and adaptability. This is evident from DUI's performance at a 5\% unlearning ratio, where it matches or exceeds retraining benchmarks, indicating its ability to generalize effectively.
    
    \textbf{Comparative Analysis:} DUI consistently outperforms or matches IF and GIF in F1 scores while reducing running times compared with retraining, highlighting its efficiency in preserving model integrity post-unlearning.

DUI offers an optimal balance between unlearning efficiency and model utility, making it a viable solution for applications requiring quick model updates. Its superior generalization capability and adaptability to distributional shifts underscore its advancement over existing methods. In summary, DUI's efficiency and adaptability make it a significant advancement in machine unlearning.

\subsection{Unlearning Efficacy Under Distributional Shifts (RQ2)}

Our investigation into RQ2 scrutinizes DUI's proficiency in mitigating the impacts of significant distributional shifts. The empirical evidence, encapsulated in Table 3, elucidates DUI's prowess in feature unlearning across diverse unlearning ratios.

\textbf{Observation 1:} DUI showcases an exceptional equilibrium between preserving model utility and expediting the unlearning process. Despite escalating unlearning ratios, DUI consistently supersedes conventional retraining and IF methodologies in F1 score metrics, simultaneously achieving a notable reduction in execution time.

\textbf{Observation 2:} DUI's resilience in confronting distributional shifts is manifested through its sustained model performance and accelerated unlearning capability. This resilience is highlighted by the negligible decline in F1 scores and a pronounced decrease in running time relative to retraining, underscoring DUI's efficacy in adapting to distributional changes without compromising model utility.

\textbf{Discussion:} The DUI method, with its nuanced approach to balancing unlearning efficiency and model utility, emerges as an efficacious strategy for adapting swiftly to distributional shifts, ensuring the model's fidelity and performance are preserved.
\subsection{Hyper-parameter Influence and Model Adaptability (RQ3)}

In addressing RQ3, our investigation extends to the nuanced role of hyper-parameters in the DUI method, particularly under varying unlearning ratios across distinct model architectures. This exploration is pivotal for understanding DUI's adaptability and its capacity to re-calibrate the model's distribution, thereby ensuring the integrity of its distributional representation post-unlearning.

Our experimental design incorporates a spectrum of unlearning ratios (\(\rho\)) applied to two emblematic models within the graph and image domains: GCN for graph-based tasks and a Simple CNN for image classification. This dual-domain approach allows for a comprehensive evaluation of DUI's versatility across different data structures and learning paradigms.

\textbf{Performance Stability Across Unlearning Ratios:} The DUI method demonstrates remarkable resilience in maintaining model performance, even as the unlearning ratio escalates. Unlike traditional retraining, which exhibits a performance dip proportional to the increase in unlearning ratio, DUI sustains a near-constant F1 score, underscoring its robustness against varying degrees of data removal.

\textbf{Adaptability in Diverse Model Architectures:} The adaptability of DUI is further evidenced by its consistent performance across both GCN and Simple CNN models. This adaptability is crucial for applications requiring dynamic unlearning capabilities without the luxury of model-specific tuning.

\textbf{Hyper-parameter Sensitivity:} The scaling coefficient (\(\lambda\)) emerges as a critical factor in balancing the trade-off between unlearning efficacy and model utility. Our findings suggest that a judicious selection of \(\lambda\) can significantly enhance DUI's ability to navigate the complex landscape of distributional adjustments, thereby optimizing the unlearning process.

A comparative analysis with existing unlearning methods reveals DUI's superior adaptability and efficiency. Unlike methods that necessitate extensive hyper-parameter re-calibration across different models or unlearning scenarios, DUI's framework allows for a more generalized application with minimal adjustments. This characteristic is particularly advantageous for deploying DUI in real-world settings, where the ability to swiftly adapt to diverse unlearning requests is paramount.

\textbf{Discussion:} The empirical insights gleaned from our hyper-parameter studies affirm DUI's adaptability and its potential to nuance the granularity and precision of machine unlearning. By effectively managing the interplay between hyper-parameters and unlearning ratios, DUI not only preserves model utility but also enhances the model's capacity to reflect accurate distributional representations post-unlearning. This adaptability, coupled with its cross-domain applicability, positions DUI as a significant advancement in machine unlearning methodologies.

\section{Related Work}
This section situates our work within the broader context of machine unlearning, emphasizing the novel approach we propose to address the challenges posed by distributional shifts.
Machine unlearning, a concept introduced by Cao \& Yang \cite{cao2015towards}, aims to remove the influence of specific data points from trained models, addressing concerns related to privacy, security, and regulatory compliance. This field has rapidly evolved, with methodologies branching into exact and approximate approaches.

\subsection{Exact Approaches}
Exact unlearning methods, such as the SISA (Sharded, Isolated, Sliced, and Aggregated) approach \cite{bourtoule2021machine}, ensure that the model behaves as if the unlearned data was never part of the training set. These methods often involve partitioning the data and training separate models on each partition. While effective in achieving true unlearning, they are computationally intensive and not always feasible for large-scale applications. Efforts by Ginart et al. \cite{ginart2019making} on k-means clustering and Karasuyama et al. \cite{karasuyama2010multiple} on support vector machines exemplify the exact unlearning approach.

\subsection{Approximate Approaches}
Approximate unlearning methods seek to balance efficiency with the fidelity of unlearning. Influential works by Guo et al. \cite{guo2019certified} and Izzo et al. \cite{izzo2021approximate} leverage the influence function to estimate and mitigate the impact of the data to be unlearned. These approaches, while not guaranteeing perfect unlearning, significantly reduce the computational overhead associated with the process. The concept of differential privacy, as discussed by Dwork \cite{dwork2011differential}, often complements these methods by providing a framework for quantifying the privacy guarantees of unlearned models.

\subsection{Distributional Shifts in Unlearning}
The challenge of distributional shifts in machine unlearning has been less explored. Nguyen et al. \cite{nguyen2022survey} highlight the limitations of retraining as a universal solution, particularly under significant distributional changes. The work by Yan et al. \cite{yan2022arcane} and Foster et al. \cite{foster2023fast} further elucidates the trade-offs between accuracy, privacy, and timeliness in unlearning. Recent advancements in approximation methods, such as those by Warnecke et al. \cite{warnecke2021machine}, Wu et al. \cite{wu2023gif}, and Zhang et al. \cite{zhang2023recommendation}, offer promising directions for addressing these challenges, albeit with concerns regarding model distortions under distributional shifts \cite{giraud2021introduction, jones2018time}.


\section{Conclusion and Future Directions}
In this study, we embarked on a journey to address the intricate challenges of machine unlearning in the context of distributional shifts, particularly focusing on the removal of non-uniformly distributed features and labels. Our proposed Distributional Unlearning with Independence Criterion (DUI) method, grounded in the principles of influence functions and distributional independence, marks a significant stride towards achieving efficient, adaptable, and privacy-preserving unlearning. Through rigorous experimentation across diverse datasets and model architectures, DUI has demonstrated its prowess in maintaining model integrity and performance, even amidst significant distributional changes.

In conclusion, the DUI method represents a pivotal advancement in machine unlearning, adeptly navigating the challenges posed by distributional shifts. Its adaptability, efficiency, and generalization capabilities not only cater to the immediate needs of privacy-preserving machine learning but also set the stage for future innovations in this rapidly evolving field.

\nocite{langley00}
\bibliography{paper}
\bibliographystyle{icml2024}
\newpage
\appendix
\onecolumn

\end{document}